\documentclass[sn-mathphys,Numbered]{sn-jnl}

\usepackage{graphicx}%
\usepackage{multirow}%
\usepackage{amsmath,amssymb,amsfonts}%
\usepackage{amsthm}%
\usepackage{mathrsfs}%
\usepackage[title]{appendix}%
\usepackage{xcolor}%
\usepackage{textcomp}%
\usepackage{manyfoot}%
\usepackage{booktabs}%
\usepackage{algorithm}%
\usepackage{algorithmicx}%
\usepackage{algpseudocode}%
\usepackage{listings}%
\usepackage{array}
\usepackage{tabularx}
\usepackage{caption}
\usepackage{geometry}
\theoremstyle{thmstyleone}%
\theoremstyle{thmstyletwo}%

\theoremstyle{thmstylethree}%

\begin{document}

\title[Article Title]{EgyBERT: A Large Language Model Pretrained on Egyptian Dialect Corpora}


\author*[1]{\fnm{Faisal} \sur{Qarah}}\email{fqarah@taibahu.edu.sa}



\affil*[1]{\orgdiv{Department of Computer Science, College of Computer Science and Engineering}, \orgname{Taibah University}, \orgaddress{\city{Medina}, \postcode{42353}, \country{Saudi Arabia}}}




\abstract{






This study presents EgyBERT, an Arabic language model pretrained on 10.4 GB of Egyptian dialectal texts. We evaluated EgyBERT's performance by comparing it with five other multidialect Arabic language models across 10 evaluation datasets. EgyBERT achieved the highest average F1-score of 84.25\% and an accuracy of 87.33\%, significantly outperforming all other comparative models, with MARBERTv2 as the second best model achieving an F1-score 83.68\% and an accuracy 87.19\%. Additionally, we introduce two novel Egyptian dialectal corpora: the Egyptian Tweets Corpus (ETC), containing over 34.33 million tweets (24.89 million sentences) amounting to 2.5 GB of text, and the Egyptian Forums Corpus (EFC), comprising over 44.42 million sentences (7.9 GB of text) collected from various Egyptian online forums. Both corpora are used in pretraining the new model, and they are the largest Egyptian dialectal corpora to date reported in the literature. Furthermore, this is the first study to evaluate the performance of various language models on Egyptian dialect datasets, revealing significant differences in performance that highlight the need for more dialect-specific models. The results confirm the effectiveness of EgyBERT model in processing and analyzing Arabic text expressed in Egyptian dialect, surpassing other language models included in the study. EgyBERT model is publicly available on \url{https://huggingface.co/faisalq/EgyBERT}. }




\maketitle

\section{Related Work}\label{sec2}

\subsection{Large-Scale Egyptian Dialectal Corpora}

The field of Arabic Natural Language Processing (NLP) has traditionally focused on Modern Standard Arabic (MSA), the standard form of written Arabic. However, in recent years, researchers have increasingly turned their attention to dialectal Arabic (DA), the informal spoken varieties of the language, due to its widespread use on social media platforms and instant messaging applications. One significant aspect of these efforts is the development of large-scale multidialectal Arabic corpora in several studies \cite{diab2010colaba, zaidan2011arabic, al2017arasenti, mubarak2014using} to address the lack of such corpora and to advance the field of dialectal Arabic NLP.


In addition to multidialectal Arabic corpora, numerous studies have focused on monodialect corpora specific to Egyptian Arabic. Al-Sabbagh and Girju introduced YADAC \cite{al2012yadac}, a multi-genre Dialectal Arabic corpus that concentrates on Egyptian Arabic. This corpus, collected from various social media and online knowledge market platforms targeting Arabic speakers, underwent a filtering process to meet specific criteria. The finalized corpus comprises 6 million wordform tokens and 457K wordform types, with a distribution of 41\% from online knowledge market services, 32\% from microblogs, and 27\% from blogs and forums. Additionally, the authors developed a related corpus, The Arabic Corpus for Egyptian Tweets \cite{al2012supervised}, which originated as a subset of YADAC. This new corpus contains 22,834 tweets, amounting to 423,691 tokens and 70,163 types, and has been manually annotated for Part-of-Speech (POS) NLP tasks.

Hussein et al. introduced the Egyptian Dialect Gender Annotated Dataset (EDGAD) \cite{hussein2019gender}. This corpus comprises 70,000 tweets per gender from 140 active Egyptian user accounts based in Egypt. For the gender identification task in their study, they employed several machine learning models, with the best-performing model achieving an accuracy of 87.6\%. Similarly, Kora and Mohammed introduced two substantial corpora: the Corpus of Arabic Egyptian Tweets \cite{DVN/LBXV9O_2019}, containing 40K tweets, and the Arabic Egyptian Corpus 2 \cite{kora2023enhanced}, which includes 10K tweets. Both corpora are annotated with sentiments labeled as either positive or negative. The authors utilized these corpora, along with additional datasets, for a sentiment analysis task employing various machine learning and deep learning models. The highest accuracy score, 93.2\%, was achieved using their proposed meta-ensemble model.

Tarmom et al. \cite{tarmom2020compression} developed two monodialectal Arabic corpora designed to detect code-switching in Arabic text: the Saudi Dialect Corpus (SDC) and the Egyptian Dialect Corpus (EDC). Both corpora are collected from Facebook and Twitter, each of these corpora comprises more than 210K words, with a total data size of 2MB per corpus.

Fashwan and Alansary developed the Morphologically Annotated Corpus for Egyptian Arabic \cite{fashwan2021morphologically}, which contains 527K words covering different genres and collected from various online sources including social media and books. The corpus features 239K words that were manually annotated based on their context, with each annotated word associated with information such as original word form, tag, glossary, gender, and conventional lemma and word form. This corpus was used to build CALIMA, a morphological analyzer for the Egyptian dialect  \cite{fashwan2022developing}, which achieved a recall score of 82.1\% in a part-of-speech (POS) tagging task.

The largest Egyptian dialect corpus reported in the literature, the Egyptian Dialect Monolingual Corpus, is introduced by Faheem et al. \cite{faheem2024improving}, containing over 15 million sentences that were collected from various online platforms, including the social networking site Fatakat, Facebook, and Twitter, with each sentence ranging from five to 50 words. Additionally, the authors developed two more corpora: the Standard Arabic monolingual corpus with 20 million sentences gathered from Wikipedia and online newspapers, and the parallel corpus featuring 40,000 Egyptian colloquial sentences manually translated into MSA by experts. These corpora were utilized to enhance Egyptian-MSA machine translation tasks using LSTM-based models.

The discussion above highlights the need for further contributions to the field of Egyptian dialect corpora. The current corpora are limited in size and are insufficient for pretraining large language models like BERT, which require substantial amount of data. The need for larger corpora is particularly critical for models targeting monodialectal text, whether for generation or analysis, as the use of dialectal Arabic continues to rise. Collecting more Egyptian dialect text is essential to enhance the performance of Arabic language models for tasks involving such dialect-specific text. In summary, despite the extensive literature on Egyptian dialect corpora, there remains a significant gap in size and diversity. Egyptian dialect corpora are still insufficient and require further contributions. To address this, we propose two new Egyptian dialectal corpora specifically designed for pretraining large language models. These additions aim to advance the field of Egyptian dialectal NLP significantly.

\subsection{Multidialect Arabic LLM}

The rise of Arabic language models such as AraBERT \cite{antoun2020arabert}, ArabicBERT \cite{safaya2020kuisail}, AraGPT2 \cite{antoun2021aragpt2}, and ARBERT \cite{abdul2020arbert} has significantly advanced the field of Arabic NLP, especially for tasks related to Modern Standard Arabic (MSA). However, in experiments on dialectal Arabic (DA) tasks, these models have failed to achieve similar performance. Given these findings, numerous researchers have been encouraged to develop multidialectal and monodialectal models designed for tackling DA tasks.

Large language models focused on MSA have made significant impact in the field of Arabic NLP. Which enable researchers to achieve uprecedentent results in a variety of tasks such as text classification, named entity recognition NER, translation, and text summarization which usually process MSA-based datasets. However, domain-specific large language models are crucial for achieving higher accuracy in specialized fields by training on texts from specific domains like legal, biomedical, or Arabic poetry, these models can better process and analyze content that meets the text nuanced features of those fields. ... Dialectal Arabic ... .

Large language models focused on MSA \cite{antoun2020arabert, safaya2020kuisail, antoun2021aragpt2, elmadany2022arat5, eddine2022arabart} have made significant impacts in the field of Arabic NLP, enabling researchers to achieve unprecedented results in a variety of tasks such as text classification, named entity recognition (NER), translation, and text summarization, which usually process MSA-based datasets. However, domain-specific large language models are crucial for achieving higher accuracy in specialized fields. By training on texts from specific domains like legal \cite{al2022aralegal}, biomedical \cite{boudjellal2021abioner}, or Arabic poetry \cite{qarah2024arapoembert}, these models can better process and analyze content that meets the nuanced features of those fields. Similarly, the development of large language models for dialectal Arabic is essential to effectively handle and interpret the informal form of the language spoken in everyday life.

Antoun et al. introduced AraBERT \cite{antoun2020arabert}, the first Arabic monolingual language model, pre-trained on 22 GB of MSA text acquired from the OSIAN \cite{osian2019zeroual} and the 1.5B Arabic Words Corpus \cite{1.5bwords}. Initially, they released multiple variants: AraBERTv1, which utilized the Farasa segmenter during text preparation, and AraBERTv0.1, which did not use any segmenter. Subsequently, they developed AraBERTv0.2, pre-trained on a significantly larger dataset of 77 GB, including text from publicly available corpora \cite{oscar2019asynchronous, osian2019zeroual, 1.5bwords}, Arabic Wikipedia, and news articles. Additionally, the authors introduced an extended variant named "AraBERTv0.2-Twitter," which was further pre-trained on 60M DA tweets.

The first language model pretrained on a combination of DA and MSA texts is QARiB, introduced by Chowdhury et al. \cite{chowdhury2020qarib}. This model's pre-training data was a mix of MSA-based corpora \cite{parker2011arabic, 1.5bwords, lison2016opensubtitles2016} and 420 million tweets, altogether amounting to 97 GB of text. Despite the relatively small portion of the pre-training data being in DA, the model has achieved higher results compared to AraBERT in DA and MSA text classification tasks.

Abdul-Mageed et al. \cite{abdul2020arbert} introduced two new BERT-based models: ARBERT and MARBERT. ARBERT was pretrained on 61 GB of MSA text from Arabic Wikipedia, online free books, and several publicly available Arabic corpora. In contrast, MARBERT was pretrained on a 128 GB dataset of DA-based texts, which included one billion Arabic tweets randomly selected from a pool of six billion. MARBERT has achieved state-of-the-art results on the majority of tasks, outperforming AraBERT and other multilingual models, particularly since most of the evaluation datasets primarily contain DA texts. Furthermore, Elmadany et al. introduced ARBERTv2 and MARBERTv2 \cite{elmadany2023orca}, enhanced variants pre-trained on larger corpora and for longer durations than their predecessors. These new models achieved unprecedented results when evaluated on the ORCA benchmark. Similarly, Inoue et al. proposed the CAMeLBERT series \cite{inoue2021camelbert}, a collection of BERT-based models pre-trained on different types of Arabic text. CAMeLBERT-MSA is pretrained on MSA corpora, CAMeLBERT-DA on DA corpora, and CAMeLBERT-CA on classical Arabic (CA) texts. Additionally, a mixed variant that was pre-trained on a combination of all the datasets used in the first three models, totaling 167 GB of text. The dataset employed in pretraining CAMeLBERT-DA includes 54 GB of DA text compiled from 28 different Arabic dialectal corpora.

In addition to multidialect language models, several studies have proposed monodialectal Arabic language models, including DziriBERT \cite{abdaoui2021dziribert} for the Algerian dialect, TunBERT \cite{haddad2023tunbert} for the Tunisian dialect, and DarijaBERT \cite{gaanoun2024darijabert} for the Moroccan dialect, and SaudiBERT \cite{qarah2024saudibert} for the Saudi dialect. These models have shown promising results when evaluated on dialect-specific datasets, outperforming other language models.



While there has been considerable progress in Arabic NLP, particularly with MSA and multidialectal Arabic, the Egyptian dialect still lacks specialized models. Egyptian Arabic, one of the most widely spoken Arabic dialects, is commonly used in media, entertainment, and daily conversations, making it important to have a language model specifically designed for it. A model pre-trained on a large corpus of Egyptian dialect would significantly enhance performance on tasks specific to the this dialect and fill a substantial gap in current modeling capabilities. In this study, all multidialect language models have been included in the analysis as comparative models. These include: AraBERTv02-Twitter, QARiB, CAMeLBERT-DA, MARBERTv1, and MARBERTv2.

\section{Methodology}\label{sec4}

In this section we elaborate on the architecture and pretraining procedure of EgyBERT, as well as the new corpora used to pretrain the model.

\subsection{Egyptian Corpora}
In this study, we have created two new corpora focused on Egyptian dialectal texts: the Egyptian Tweets Corpus (ETC) and the Egyptian Forums Corpus (EFC). These corpora are intended for pre-training EgyBERT and other large language models targeting the Egyptian dialect. To our knowledge, they represent the largest Egyptian dialectal corpora reported in the literature.

\subsubsection{Egyptian Tweets Corpus}


The Egyptian Tweets Corpus (ETC)\footnote{\url{https://huggingface.co/datasets/faisalq/ETC}} was acquired from a large in-house dataset containing over 800 million Arabic tweets. We extracted the ETC tweets  based on the 'location' metadata field, where users typically include the names of their city or region. To identify tweets from Egyptian users, we conducted a thorough search using both Arabic and English terms associated with Egyptian cities and regions, by examining if the 'location' field contained any of the 52 Egyptian-related terms we had manually compiled. The text in the 'location' field required extensive cleaning and preprocessing to standardize the varied writing styles used by different users. This preprocessing included removing consecutive letter repetitions, diacritics, symbols, punctuation, and emojis (excluding the Egyptian flag emoji), as well as eliminating Arabic letter extensions (Tatweel). We also normalized the Arabic text by converting various letter forms to a single standard form and transformed accented English characters to their unaccented equivalents. These measures enabled us to retrieve a total of 34,336,598 tweets.

Before using the ETC corpus in pretraining EgyBERT model, we applied several preprocessing procedures to the corpus text. These included the removal of URLs, user mentions, hashtags, and numbers larger than seven digits. Additionally, the repetition of letters was capped at five occurrences, while other characters and emojis were limited to four. Tweets with fewer than three words or those containing more than 50\% English text were also removed. Following these steps, all duplicate texts were also eliminated. Algorithm \ref{alg1} outlines the preprocessing procedures applied to the ETC corpus before its use in pretraining the model. After preprocessing, the final size of the ETC corpus is 2.5 GB, containing 270,079,892 words and 24,892,951 sentences.

\begin{algorithm}
\caption{Text Preprocessing for ETC Corpus}
\label{alg1}
\begin{algorithmic}[1]

\Function{PreprocessText}{$tweets$}
    \State Initialize an empty list $processedTweets$
    \For{each $tweet$ in $tweets$}
        \State $tweet \gets$ \Call{RemoveURLsMentionsHashtags}{$tweet$}
        \State $tweet \gets$ \Call{RemoveNewlinesAndWhitespace}{$tweet$}
        \State $tweet \gets$ \Call{RemoveLargeNumbers}{$tweet$}
        \State $tweet \gets$ \Call{LimitLetterRepetition}{$tweet$}
        \State $tweet \gets$ \Call{LimitCharacterRepetition}{$tweet$}
        
        \If{\Call{WordCount}{$tweet$} \textless{} 3}
            \State \textbf{continue}
        \EndIf
        \If{\Call{IsMajorityEnglish}{$tweet$}}
            \State \textbf{continue}
        \EndIf
        \State Append $tweet$ to $processedTweets$
    \EndFor
    \State $processedTweets \gets$ \Call{RemoveDuplicates}{$processedTweets$}
    \State \Return $processedTweets$
\EndFunction

\end{algorithmic}
\end{algorithm}

\begin{figure}[]
    \centering
    \includegraphics[width=0.7\linewidth]{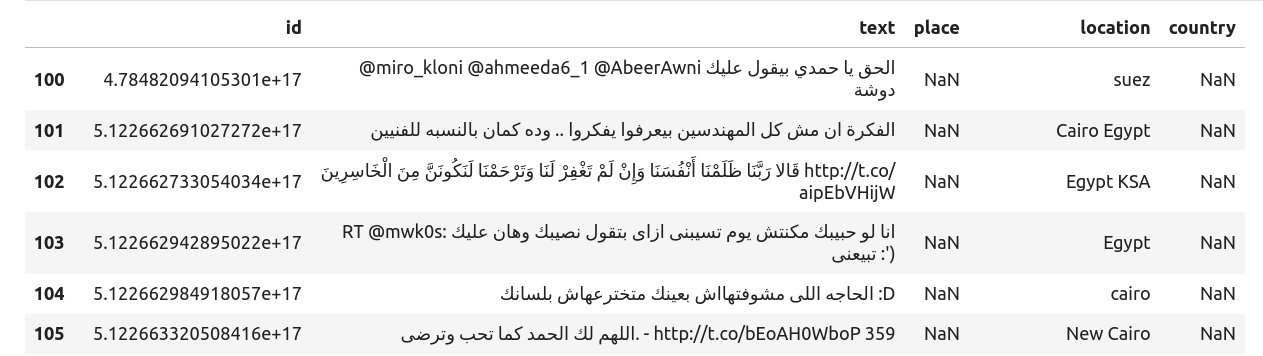}
        \vspace{0.2cm} 
    \caption{Sample of ETC corpus.}
    \label{fig4}
\end{figure}

The ETC corpus is publicly available, but in compliance with Twitter's terms of service we have only released the tweet IDs, which can be used to hydrate the tweets text using Twitter API. Figure \ref{fig4} shows a sample of ETC corpus.

{}





\subsubsection{Egyptian Forums Corpus}

The Egyptian Forums Corpus (EFC) is the second corpus developed in this study following the ETC. It was compiled by collecting texts from four different Egyptian online forums: Fmisr\footnote{\url{https://fmisr.com/}}, Kooora\footnote{\url{https://www.kooora.com/}}, Almatareed\footnote{\url{https://www.almatareed.org/vb/}}, and Banatmasr\footnote{\url{https://banatmasr.net/}}. The process of acquiring the text began by downloading all the HTML files from specific sections of each forum, covering a wide range of genres including sports, health, traveling, technology, politics, and religion. We then extracted the text from each file locally using the BeautifulSoup API \cite{richardson2007beautiful}.

Following the collection of the desired texts, we applied the same preprocessing procedures to the EFC corpus that were used for the ETC. This included removing URLs, email addresses, and large numbers; limiting the repetition of letters to five occurrences and other characters to four; and removing texts with fewer than three words or more than 50\% of their content in English. Additionally, we eliminated BBCode tags, HTML tags, percent-encoded sequences, and all duplicate texts. Algorithm \ref{alg2} presents the preprocessing steps applied to the EFC corpus.

\begin{algorithm}
\caption{Text Preprocessing for EFC Corpus}
\label{alg2}
\begin{algorithmic}[1]

\Function{PreprocessText}{$texts$}
    \State Initialize an empty list $processedTexts$
    \For{each $text$ in $texts$}
        \State $text \gets$ \Call{RemoveURLs}{$text$}
        \State $text \gets$ \Call{RemoveEmails}{$text$}
        \State $text \gets$ \Call{RemoveNewlinesAndWhitespace}{$text$}
        \State $text \gets$ \Call{RemoveLargeNumbers}{$text$}
        \State $text \gets$ \Call{LimitLetterRepetition}{$text$}
        \State $text \gets$ \Call{LimitCharacterRepetition}{$text$}
        \State $text \gets$ \Call{RemoveBBCode}{$text$}
        \State $text \gets$ \Call{RemoveHTMLTags}{$text$}
        \State $text \gets$ \Call{RemovePercentEncodedSequences}{$text$}
        
        \If{\Call{WordCount}{$text$} \textless{} 3}
            \State \textbf{continue}
        \EndIf
        \If{\Call{IsMajorityEnglish}{$text$}}
            \State \textbf{continue}
        \EndIf
        \State Append $text$ to $processedTexts$
    \EndFor
    \State $processedTexts \gets$ \Call{RemoveDuplicates}{$processedTexts$}
    \State \Return $processedTexts$
\EndFunction

\end{algorithmic}
\end{algorithm}

\begin{figure}[]
    \centering
    \includegraphics[width=0.8\linewidth]{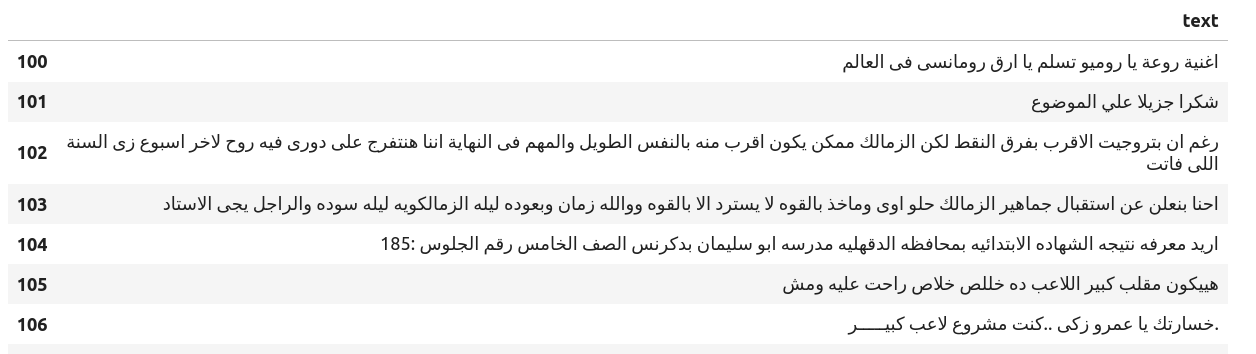}
    \caption{Sample of EFC corpus.}
    \label{fig5}
\end{figure}

The final size of the EFC corpus is 7.9 GB, containing 805,807,960 words and 44,423,062 sentences. The complete EFC corpus will not be made public. However, we are releasing a subset called EFC-mini\footnote{\url{https://huggingface.co/datasets/faisalq/EFC-mini}}, which represents 20\% of the full corpus that was randomly sampled. EFC-mini contains 201,286,815 words and 11,109,033 sentences. Figure \ref{fig5} shows a sample of EFC corpus.








{}


\subsection{EgyBERT Pretraining Procedures}

The proposed model follows the same architecture as the original BERT model containing 12 encoder layers, 12 attention heads per layer, a hidden layer size of 768 units, the vocabulary size of is set to 75,000 wordpieces. For tokenizing the pretraining corpora, we used the SentencePiece tokenizer \cite{kudo2018sentencepiece} based on our previous work \cite{qarah2024comprehensive} that shows employing SentencePiece tokenizer with Arabic language models yield better performance. Furthermore, EgyBERT model was pretrained solely on the masked language model (MLM) training objective task, whereas 15\% of input text's tokens are masked. To optimize the training process and enhance GPU memory usage efficiency while being pretrained on a commodity GPU on a local machine, we used the mixed precision data type "FP16" for gradient computations. The model was pretrained for 24.5 epochs (10.65 million steps, 774 hours) with a maximum sequence length of 128 and a batch size of 64. We utilized the 'AdamW' optimizer with a learning rate of 5e-5, and achieved a training loss of 2.88.



\section{Experiments}\label{sec5}

In this section, we describe the datasets used in the evaluation process and the fine-tuning procedures. All experiments, including text preprocessing, model pretraining, and fine-tuning, were conducted on a local machine with an AMD Ryzen-9 7950x processor, 64GB of DDR5 memory, and two GeForce RTX 4090 GPUs, each with 24GB of memory. The software environment was set up on the Ubuntu 22.04 operating system, using CUDA 11.8 and the Huggingface transformers library \cite{wolf2020transformers} to download and fine-tune the comparative language models from the Huggingface hub.

\subsection{Fine-tuning Procedures}

In this study, we used consistent hyperparameters across all experiments when fine-tuning the models. Each model was fine-tuned with a batch size of 64, a maximum sequence length of 128, and the 'AdamW' optimizer with a learning rate of 5e-5. To optimize training efficiency, we employed the mixed precision data type "FP16" for gradient computations.

In all experiments, models were evaluated using the F1-score and Accuracy metrics. Each experiment was conducted three times, and the highest F1-score achieved by each model was reported for each task. The number of epochs and validation steps varied depending on the number of training and validation samples, as well as the complexity of each dataset. We experimented with different numbers of epochs for each task until a suitable number was found that was appropriate for all models. For each task, the dataset was split into 80\% for training and 20\% for validation, except for the iSarcasmEval dataset, which was already divided into training and validation sets by the authors. The same validation set was used to evaluate all models within the same experiment to ensure a fair comparison.

\subsection{Evaluation Tasks}

In this study, we assessed the performance of EgyBERT alongside five  multidialectal language models: AraBERTv02-Twitter\footnote{\url{https://huggingface.co/aubmindlab/bert-base-arabertv02-twitter}}, QARiB\footnote{\url{https://huggingface.co/qarib/bert-base-qarib}}, CAMeLBERT-DA\footnote{\url{https://huggingface.co/CAMeL-Lab/bert-base-arabic-camelbert-da}}, MARBERTv1\footnote{\url{https://huggingface.co/UBC-NLP/MARBERT}}, and MARBERTv2\footnote{\url{https://huggingface.co/UBC-NLP/MARBERTv2}}. Table \ref{tab0} shows the models characteristics and the pre-training datasets. 
All models were evaluated on 10 publicly available Egyptian dialectal datasets designed for various NLP tasks, such as sentiment analysis, text classification, sarcasm detection, and gender identification. 
Although some datasets required text preprocessing, but we chose to experiment with the datasets in their original form without performing any additional preprocessing to maintain the integrity of the original data and to evaluate the models' performance on raw, unprocessed text. Table \ref{tab11} provides a brief summary of the evaluation datasets included in the study.

\begin{table}[ht]
\centering
\caption{Characteristics details for all models included in the study}
\label{tab0}
\begin{tabular}{lccccc}
\toprule
\textbf{Model} & \textbf{\#Parameters} & \textbf{Tokenizer} & \textbf{Vocab-Size} & \textbf{Dataset Size} & \textbf{Dataset Type}  \\ \midrule
QARiB & 136M & WordPiece & 64K & 97GB & MSA+DA  \\
MARBERTv1 & 163M & WordPiece & 100K & 128GB & MSA+DA  \\
MARBERTv2 & 163M & WordPiece & 100K & 128GB & MSA+DA \\
CAMeLBERT-DA & 108M & WordPiece & 30K & 54GB & DA  \\
AraBERTv0.2-Twitter & 136M & WordPiece & 64K & 77GB & MSA+DA  \\

EgyBERT & 144M & SentencePiece & 75K & 10.4GB & Egyptian DA  \\
 \bottomrule
\end{tabular}
\end{table}


\begin{table}[h]
\centering
\caption{Summary of various Arabic Egyptian datasets included in the study}
\label{tab11}
\begin{tabular}{lccccp{5cm}}
\hline
\textbf{Dataset} & \textbf{Type} & \textbf{\#Labels} & \textbf{Train Size} & \textbf{Dev Size} & \textbf{Link} \\
\hline
SDC-EDC & Dialect Detection & 2 & 22.89k & 5.72k &   \url{https://github.com/TaghreedT} \\

AbusiveLanguage & Abusive Language Detection & 3 & 0.88k & 0.22k &   \url{http://alt.qcri.org/~hmubarak/offensive/TweetCorpus2017_English_MSA_ARA.zip} \\

iSarcasmEval (Nile dialect) & Sarcasm Detection & 2 & 1,29k & 0.52k &  \url{https://github.com/iabufarha/iSarcasmEval} \\

EDGAD & Gender Identification & 2 & 80k & 20k & \url{https://github.com/shery91/Egyptian-Dialect-Gender-Dataset} \\

CorpusOfArabicEgyptianTweets & Sentiment Analysis & 2 & 32k & 8k &   \url{https://dataverse.harvard.edu/dataset.xhtml?persistentId=doi:10.7910/DVN/LWPY9H} \\

ArabicEgyptianCorpus2 & Sentiment Analysis & 2 & 8k & 2k &   \url{https://dataverse.harvard.edu/dataset.xhtml?persistentId=doi:10.7910/DVN/LWPY9H} \\

EmotionalTone & Emotion Analysis & 8 & 8k & 2k &   \url{https://github.com/amrmalkhatib/Emotional-Tone} \\

ArSAS (sentiment) & Sentiment Analysis & 4 & 15.19k & 3.98k &  \url{https://huggingface.co/datasets/arbml/ArSAS} \\

ArSAS (topic) & Topic Characteristic & 3 & 15.19k & 3.98k &    \\

EgyptianCompaniesReviews & Sentiment Analysis & 3 & 32k & 8k & \url{https://github.com/zahran1234/transfer-learning} \\
\hline
\end{tabular}
\end{table}


\begin{itemize}

    \item 
    \textbf{SDC-EDC:} \\
    The Saudi Dialect Corpus (SDC) and the Egyptian Dialect Corpus (EDC) are two separate collections of texts compiled by Tarmom et al. \cite{tarmom2020compression} from Facebook and Twitter. The SDC contains 14,891 entries, and the EDC includes 13,739. We combined both corpora after labeling each one with its respective dialect to evaluate the performance of various language models in identifying and differentiating between Saudi and Egyptian dialectal texts.

\vspace{2mm}

    \item 
    \textbf{Abusive Language:} \\
    The Abusive Language dataset introduced by Mubarak et al. \cite{mubarak2017abusive} is designed to evaluate the effectiveness of various machine learning models in detecting abusive text primarily expressed in the Egyptian dialect. This dataset comprises 1,100 tweets collected from Egyptian users accounts on Twitter, which have been manually annotated into three categories: obscene, offensive (but not obscene), and clean. The distribution of the annotations is 19.1\% obscene, 40.3\% offensive, and 40.6\% clean, respectively.

\vspace{2mm}

    \item 
    \textbf{iSarcasmEval:} \\
The iSarcasmEval dataset \cite{farha2022semeval} originally contains 4,502 entries, each manually labeled for both dialect and sarcasm (either sarcastic or non-sarcastic) across five dialects: Gulf, Levant, Maghreb, MSA, and Nile. For our study, we focused on the Nile dialect, indicating Egyptian dialect, which includes 1,814 texts. Within this subset, 62\% of the texts are labeled as sarcastic and 38\% as non-sarcastic.


\vspace{2mm}

    \item 
    \textbf{EDGAD:} \\
    The Egyptian Dialect Gender Annotated Dataset (EDGAD) \cite{hussein2019gender} was introduced for the gender identification task. Initially, the dataset contained 140K tweets evenly split between genders. However, the dataset link currently includes approximately 200K tweets per gender. For our study, we reduced the dataset to the first 50K tweets for each gender, which still makes it the largest dataset used in the evaluation process.

\vspace{2mm}

     \item 
    \textbf{Corpus Of Arabic Egyptian Tweets:} \\
    Another dataset based on Twitter is the Corpus Of Arabic Egyptian Tweets \cite{DVN/LBXV9O_2019} which is designed for the sentiment analysis task. The authors collected 40K tweets in Egyptian dialect that were manually annotated to either positive or negative sentiments, with both labels evenly distributed throughout the dataset.

\vspace{2mm}

     \item 
    \textbf{Arabic Egyptian Corpus 2:} \\
    The same authors of the previous dataset have introduced an extension called the Arabic Egyptian Corpus-2 \cite{kora2023enhanced}, which consists of 10,000 tweets labeled for sentiment analysis. The dataset is evenly distributed between the two sentiments: positive and negative.

\vspace{2mm}

     \item 
    \textbf{EmotionalTone:} \\
The EmotionalTone dataset \cite{al2018emotional} contains 10,000 tweets filtered by Egypt's geo-location to ensure relevance to the Egyptian dialect. These tweets are manually labeled with eight emotions: joy, anger, sympathy, sadness, fear, surprise, love, and none. The dataset maintains a balanced distribution of emotions, with 'surprise' having the least occurrences at 1,045 tweets and 'none' being the most frequent with 1,550 tweets.

    \vspace{2mm}

     \item 
    \textbf{ArSAS:} \\
The Arabic Speech Act and Sentiment (ArSAS) dataset \cite{abdelrahim2018arsas} is a compilation of Egyptian tweets designed for various NLP tasks such as sentiment analysis, speech-act detection, and topic characteristic identification. In this study, we included the sentiment analysis and topic identification tasks. The publicly available version of the dataset contains 19,897 tweets, although the original dataset includes 21,064 tweets.
For sentiment analysis, the dataset is annotated with four different sentiments: positive, negative, neutral, and mixed. The distribution of sentiments is as follows: 7,384 tweets labeled as negative, 6,894 as neutral, 4,400 as positive, and 1,219 as mixed. Additionally, for the topic characteristics (speech type) task, the dataset categorizes tweets into one of three classes: Entity (5,810 tweets), Event (8,288 tweets), and Long-Standing (5,799 tweets).


    

    \vspace{2mm}

     \item 
    \textbf{Egyptian Companies Reviews:} \\
The Egyptian Companies Reviews dataset \cite{CompaniesReviews} is designed for sentiment analysis on reviews of 12 Egyptian companies collected from Twitter and Google Play. It consists of 40,045 entries, each labeled with one of three sentiments: negative, neutral, or positive. The sentiment distribution within the dataset is 35.45\% negative, 59.75\% positive, and 4.8\% neutral.

    

\end{itemize}

\begin{table}[h]
\centering
\caption{Evaluation results of various models on the Egyptian dialect datasets. All results are truncated to two decimal places. The average score is calculated as the unweighted average of all scores prior to truncation.}
\label{tab33}
\begin{tabular}{lcccccccccccc}
\hline

  & \multicolumn{2}{c}{\textbf{EgyBERT}} & \multicolumn{2}{c}{\textbf{MARBERTv2}} & \multicolumn{2}{c}{\textbf{MARBERTv1}} & \multicolumn{2}{c}{\textbf{AraBERT0.2}} & \multicolumn{2}{c}{\textbf{QARiB}} & \multicolumn{2}{c}{\textbf{CAMeLBERT}}\\
\hline
  \textbf{Dataset}& \textbf{Acc.} & \textbf{F1} & \textbf{Acc.} & \textbf{F1} & \textbf{Acc.} & \textbf{F1} & \textbf{Acc.} & \textbf{F1} & \textbf{Acc.} & \textbf{F1} & \textbf{Acc.} & \textbf{F1} \\
\hline
SDC-EDC & \textbf{98.09} & \textbf{98.09} & 97.88 & 97.88 & 97.41 & 97.41 & 97.58 & 97.58 & 97.79 & 97.79 & 97.27 & 97.26 \\

AbusiveLanguage & \textbf{79.54} & \textbf{81.07} & 79.09 & 79.24 & 79.09 & 78.06 & 79.09 & 78.73 & 79.09 & 79.10 & 75.90 & 76.09 \\

iSarcasmEval (Nile dialect) & \textbf{82.11} & \textbf{77.42} & 80.19 & 75.32 & 78.46 & 74.97 & 74.23 & 70.81 & 79.03 & 74.21 & 75.57 & 69.95 \\

EDGAD & \textbf{83.26} & \textbf{83.26} & \textbf{83.26} & \textbf{83.26} & 81.70 & 81.69 & 83.23 & 83.23 & 82.91 & 82.91 & 81.49 & 81.48 \\

ArabicEgyptianTweets & \textbf{91.66} & \textbf{91.65} & 91.53 & 91.53 & 90.63 & 90.63 & 91.25 & 91.24 & 91.01 & 91.01 & 89.11 & 89.11 \\

ArabicEgyptianCorpus2 & \textbf{98.80} & \textbf{98.79} & 98.70 & 98.74 & 98.20 & 98.24 & 98.60 & 98.65 & 98.10 & 98.15 & 96.50 & 96.49 \\

EmotionalTone & \textbf{79.23} & \textbf{79.40} & 78.93 & 79.14 & 78.34 & 78.31 & 77.69 & 78.00 & 77.14 & 77.39 & 75.70 & 75.90 \\

ArSAS (sentiment) & 75.62 & 66.55 & 76.90 & 66.34 & 76.38 & 65.89 & \textbf{77.08} & \textbf{67.50} & 76.25 & 66.16 & 75.35 & 65.27 \\

ArSAS (topic) & 99.54 & 99.54 & \textbf{99.57} & \textbf{99.55} & 99.54 & \textbf{99.55} & 99.27 & 99.27 & 99.22 & 99.22 & 99.19 & 99.19 \\

EgyptianCompaniesReviews & 85.50 & \textbf{66.74} & \textbf{85.80} & 65.81 & 84.56 & 65.61 & 84.29 & 66.05 & 86.27 & 65.96 & 84.79 & 65.25 \\

\hline
Avg. & \textbf{87.33} & \textbf{84.25} & 87.19 & 83.68 & 86.43 & 83.04 & 86.23 & 83.11 & 86.69 & 83.19 & 85.09 & 81.60 \\
\hline
\end{tabular}
\end{table}

\section{Results}\label{sec6}

In this section we present the evaluation results of all language models tested on the validation set of each task using 'Accuracy' and 'Macro F1-score' as evaluation metrics. Table \ref{tab33} shows that EgyBERT achieved the highest F1-score in eight out of ten tasks and the highest Accuracy score in seven tasks. However, EgyBERT was slightly behind MARBERTv2, MARBERTv1, and AraBERTv0.2-Twitter in the tasks related to ArSAS dataset, which contains a significant amount of text in MSA and other dialects, alongside the Egyptian dialect that EgyBERT is specifically designed to process efficiently.

Overall, EgyBERT has significantly outperformed all other models with an average Accuracy of 87.33\% and an average F1-score of 84.25\%, despite being trained on a corpus significantly smaller than those used by the other models, which were pretrained on text collections 5 to 12 times larger. MARBERTv2 was the second best model which has achieved average scores of 87.19\% for Accuracy and 83.68\% for F1-score respectively.
Followed by MARBERTv1, AraBERTv0.2-Twitter, and QARiB which performed similarly, achieving very close results.
Finally, CAMeLBERT-DA performed the worst achieving significantly lower results compared to the other models included in the study, which makes it the least preferable option for tasks involving datasets that primarily contain Egyptian dialectal text.

\section{Conclusion}\label{sec7}

In this study, we introduced EgyBERT, the first language model specifically designed and pre-trained for the Arabic Egyptian dialect. We leveraged two extensive, novel corpora: the Egyptian Tweets Corpus (ETC) and the Egyptian Forums Corpus (EFC). These are the largest Egyptian dialectal corpora to date, compiled from various social media platforms.

Our experiments show that EgyBERT outperforms existing multidialectal Arabic language models on the majority of tasks, with an average Accuracy of 87.33\% and an average F1-score of 84.25\% when evaluated on 10 several NLP tasks in Egyptian dialect, showing the effectiveness of using a domain-specific model such as EgyBERT.

The corpora and the language model introduced in this paper will serve as valuable resources for future research across multiple fields including linguistics, artificial intelligence, language processing, and cultural studies. Furthermore, the presented corpora can be incorporated into a larger pretraining dataset for future, more expansive models that necessitate trillions of tokens.


\backmatter








\section*{Declarations}




\subsection{Conflict of interest}
The author declares no conflict of interest.

\subsection{Availability of data and materials}
EgyBERT model is publicly available on \url{https://huggingface.co/faisalq/EgyBERT}. ETC dataset is available on \url{ https://huggingface.co/faisalq/ETC}, and EFC-mini is available on \url{ https://huggingface.co/faisalq/EFC-mini}. 

\subsection{Code availability}
The code and the results are available on \url{https://github.com/FaisalQarah/EgyBERT}.

\bibliography{sn-bibliography}

\end{document}